\pdfoutput=1
\documentclass[11pt,a4paper]{article}
\usepackage[hyperref]{acl2019}
\usepackage{times}
\usepackage{latexsym}

\usepackage{url}
\usepackage{xspace}

\aclfinalcopy 

\usepackage{tikz}
\usetikzlibrary{positioning,arrows.meta,patterns}
\tikzset{execute at begin node=\strut}
\tikzset{center coordinate/.style={
		execute at end picture={\path
			([rotate around={180:#1}]perpendicular cs:
			horizontal line through={#1},
			vertical line through={(current bounding box.east)})
			([rotate around={180:#1}]perpendicular cs:
			horizontal line through={#1},
			vertical line through={(current bounding box.west)});
		}
}}
\tikzstyle{defmeta}=[on grid, style=thick, inner sep=0pt]
\newcommand{\defxy}[3]{\tikzstyle{#1}=[defmeta, node distance=#3 mm and #2 mm, x=#2 mm, y=#3 mm]}
\defxy{default}{25}{20}
\defxy{defnarrow}{22}{20}
\defxy{defwide}{60}{17}
\defxy{defshort}{25}{15}
\defxy{defcompact}{22}{15}
\tikzstyle{scopetree}=[on grid, node distance = 10mm and 12mm, inner sep=.5mm]
\tikzstyle{scopetreewide}=[on grid, node distance = 12mm and 24mm]
\tikzstyle{scopetreewidedeep}=[on grid, node distance = 18mm and 24mm]
\tikzstyle{node}=[circle, draw, minimum size=10mm]
\tikzstyle{ent}=[node, fill=porange]
\tikzstyle{val}=[node, fill=plila]
\tikzstyle{tok}=[node, fill=grau]
\tikzstyle{sit}=[node, fill=qorange]
\tikzstyle{func}=[node, fill=qlila]
\tikzstyle{bin}=[rectangle, draw, minimum size=2.1mm, inner sep=-2mm, font=\tiny]
\tikzstyle{comp}=[bin, fill=orange]
\tikzstyle{sig}=[bin, fill=lila]
\tikzstyle{link1}=[out=106, in=74, looseness=.5]
\tikzstyle{link2}=[out=102, in=78, looseness=.6]
\tikzstyle{link3}=[out=98, in=82, looseness=.69]
\tikzstyle{link4}=[out=94, in=86, looseness=.77]
\tikzstyle{link5}=[out=90, in=90, looseness=.84]
\tikzstyle{dir}=[-{Straight Barb[angle=60:2.2mm]}, shorten > =0.2mm]
\tikzstyle{dircomp}=[{Straight Barb[angle=90:2.2mm]}-, shorten < =0.2mm]
\tikzstyle{dirextra}=[shorten < =0.5mm]
\tikzstyle{inf}=[densely dotted, {Straight Barb[angle=60:2.2mm]}-, shorten < =0.2mm]
\tikzstyle{scope}=[dotted, lila, {Straight Barb[angle=60:2.2mm]}-, shorten < =0.2mm]
\newlength{\compsep}
\newlength{\scopesep}
\newlength{\funcsep}
\definecolor{lila}{RGB}{159,114,207}
\definecolor{orange}{RGB}{230,130,50}
\colorlet{plila}{lila!60!white}
\colorlet{qlila}{lila!20!white}
\colorlet{porange}{orange!60!white}
\colorlet{qorange}{orange!20!white}
\definecolor{braun}{tHsb}{27,1,.5}
\definecolor{violett}{tHsb}{269,1,.5}
\definecolor{hellviolett}{tHsb}{269,1,.8}
\definecolor{grau}{RGB}{192,192,192}

\usepackage{tabularx} 
\newcolumntype{C}{>{\centering\arraybackslash}X} 

\usepackage{hhline}

\usepackage{amssymb}
\usepackage{amsmath}	
\usepackage{nicefrac}
\usepackage{xfrac}
\usepackage{bbm}

\newcommand{\eq}{\!=\!}  
\newcommand{\mn}{\!-\!}

\newcommand{\rd}[1]{\left(#1\right)}  
\newcommand{\sq}[1]{\left[#1\right]}  
\newcommand{\cl}[1]{\left\{#1\right\}}  

\newcommand{\E}[1]{\mathbb{E}_{#1}}  
\newcommand{\Eo}[2]{\E{#1}\sq{#2}}
\renewcommand{\Pr}{\mathbb{P}}  
\newcommand{\Po}[1]{\Pr\rd{#1}}  
\newcommand{\Pc}[2]{\Po{#1\,\middle|\,#2}}  
\newcommand{\R}{\mathcal{R}(v)}  
\newcommand{\B}{\mathcal{B}(v)}  

\usepackage[nameinlink,capitalize]{cleveref}  
\crefname{equation}{}{}
\crefformat{section}{#2\S#1#3}
\crefformat{subsection}{#2\S#1#3}
\crefmultiformat{section}{#2\S#1#3}{ and~#2\S#1#3}{, #2\S#1#3}{, and~#2\S#1#3}
\crefmultiformat{subsection}{#2\S#1#3}{ and~#2\S#1#3}{, #2\S#1#3}{, and~#2\S#1#3}
\crefrangeformat{section}{#3\S#1#4--#5#2#6}
\crefrangeformat{subsection}{#3\S#1#4--#5#2#6}

\newcommand{\seeref}[1]{(see~\cref{#1})}
\newcommand{\citex}[3][]{\citep[#2:][#1]{#3}}  
\newcommand{\citeg}[2][]{\citex[#1]{for example}{#2}}  
\newcommand{\sortbib}[1]{}

\defcitealias{emerson2017b}{E\&C}
\newcommand{\EC}{\citetalias{emerson2017b}\xspace}

\makeatletter
\def\new@fontshape{}
\makeatother
\let\oldexp\exp
\usepackage{gb4e}
\noautomath
\let\exp\oldexp
\crefformat{xnumi}{#2(#1)#3}
\crefmultiformat{xnumi}{#2(#1)#3}{ and~#2(#1)#3}{, #2(#1)#3}{, and~#2(#1)#3}
\crefrangeformat{xnumi}{#3(#1)#4--#5(#2)#6}
\crefformat{xnumii}{#2(#1)#3}
\crefformat{xnumiii}{#2(#1)#3}
\crefformat{xnumiv}{#2(#1)#3}

\title{Linguists Who Use Probabilistic Models Love Them:\\
	Quantification in Functional Distributional Semantics}

\author{Guy Emerson \\
	Department of Computer Science and Technology \\
	University of Cambridge \\
	\texttt{gete2@cam.ac.uk}}

\date{}

\begin{document}
\maketitle
\begin{abstract}
	\vspace*{-.5mm}
	Functional Distributional Semantics
	provides a computationally tractable framework
	for learning truth-conditional semantics
	from a corpus.
	Previous work in this framework
	has provided a probabilistic version of first-order logic,
	recasting quantification as Bayesian inference.
	In this paper, I show how
	the previous formulation gives trivial truth values
	when a precise quantifier is used with vague predicates.
	I propose an improved account,
	avoiding this problem by treating a vague predicate
	as a distribution over precise predicates.
	I connect this account to recent work
	in the Rational Speech Acts framework
	on modelling generic quantification,
	and I extend this to modelling donkey sentences.
	Finally, I explain how the generic quantifier
	can be both pragmatically complex
	and yet computationally simpler than precise quantifiers.
	\vspace*{1mm}
\end{abstract}

\section{Introduction}

Model-theoretic semantics defines meaning in terms of \textit{truth},
relative to \textit{model structures}.
In the simplest case, a model structure consists of a set of \textit{individuals}
(also called \textit{entities}).
The meaning of a content word is a \textit{predicate},
formalised as a \textit{truth-conditional function}
which maps individuals to \textit{truth values}
(either \textit{truth} or \textit{falsehood}).
Because of this precisely defined notion of truth,
model theory naturally supports logic,
and has become a prominent approach to formal semantics.
For detailed expositions, see:
\citet{cann1993sem,allan2001sem,kamp2013sem}.

Mainstream approaches to distributional semantics
represent the meaning of a word as a vector
(for example: \citealp{turney2010vector,mikolov2013vector};
for an overview, see: \citealp{emerson2020goals}).
In contrast, Functional Distributional Semantics
represents the meaning of a word as a truth-conditional function
\citep{emerson2016,emerson2018}.
It is therefore a promising framework
for automatically learning truth-conditional semantics from large datasets.

In previous work
(\citealp{emerson2017b}, \S3.5, henceforth \EC),
I sketched how this approach can be extended
with a probabilistic version of first-order logic,
where quantifiers are interpreted in terms of conditional probabilities.
I summarise this approach in~\cref{sec:gen-quant,sec:func-dist-sem}.

There are four main contributions of this paper.
In~\cref{sec:vague-quant:trivial},
I first point out a problem with my previous approach.
Quantifiers like \textit{every} and \textit{some}
are treated as precise, but predicates are vague.
This leads to trivial truth values,
with \textit{every} trivially false, and \textit{some} trivially true.

Secondly, I show
in~\cref{sec:vague-quant:prob-tree,sec:vague-quant:dist-prec,sec:vague-quant:prob-tree-soft}
how this problem can fixed by treating
a vague predicate as a distribution over precise predicates.

Thirdly, in~\cref{sec:generic}
I look at vague quantifiers and generic sentences,
which present a challenge for classical (non-probabilistic) theories.
I build on \citet{tessler2019generic}'s
account of generics using Rational Speech Acts,
a Bayesian approach to pragmatics \citep{frank2012rsa}.
I show how generic quantification
is computationally simpler than classical quantification,
consistent with evidence that generics are a ``default'' mode of processing
\citeg{leslie2008generic,gelman2015generic}.

Finally, I show in~\cref{sec:donkey}
how this probabilistic approach can provide an account of donkey sentences,
another challenge for classical theories.
In particular, I consider generic donkey sentences,
which are doubly challenging,
and which provide counter-examples
to the claim that donkey pronouns are associated with universal quantifiers.

Taking the above together, in this paper
I show how a probabilistic first-order logic
can be associated with a neural network model for distributional semantics,
in a way that sheds light on long-standing problems in formal semantics.

\section{Generalised Quantifiers}
\label{sec:gen-quant}

\citet{partee2012history} recounts how quantifiers have played an important role
in the development of model-theoretic semantics,
seeing a major breakthrough with \citet{montague1973model}'s work,
and culminating in the theory of \textit{generalised quantifiers}
\citep{barwise1981quant,vanbenthem1984quant}.

Ultimately, model theory requires quantifiers to give truth values to propositions.
An example of a logical proposition is given in \cref{fig:sem-comp-fol},
with a quantifier for each logical variable.
This also assumes a neo-Davidsonian approach to event semantics
\cite{davidson1967event,parsons1990event}.

Equivalently, we can represent a logical proposition as a \textit{scope tree},
as in \cref{fig:scope-tree}.
The truth of the scope tree can be calculated
by working bottom-up through the tree.
The leaves of the tree are logical expressions with free variables.
They can be assigned truth values
if each variable is fixed as an individual in the model structure.
To assign a truth value to the whole proposition,
we work up through the tree,
quantifying the variables one at at time.
Once we reach the root, all variables have been quantified,
and we are left with a truth value.

Each quantifier is a non-terminal node with two children
-- its \textit{restriction} (on the left)
and its \textit{body} (on the right).
It quantifies exactly one variable, called its \textit{bound variable}.
Each node also has \textit{free variables}.
For each leaf, its free variables are exactly the variables appearing in the logical expression.
For each quantifier, its free variables are
the union of the free variables of its restriction and body,
minus its own bound variable.
For a well-formed scope tree, the root has no free variables.
Each node in the tree defines a truth value,
given a fixed value for each free variable.

The truth value for a quantifier node
is defined based on its restriction and body.
Given values for the quantifier's free variables,
the restriction and body only depend on the quantifier's bound variable.
The restriction and body therefore each define a set of individuals in the model structure
-- the individuals for which the restriction is true,
and the individuals for which the body is true.
We can write these as $\R$~and~$\B$, respectively,
where $v$~denotes the values of all free variables.

Generalised quantifier theory says that
a quantifier's truth value only depends on two quantities:
the cardinality of the restriction~$|\R|$, and
the cardinality of the intersection of the restriction and body $|\R\cap\B|$.
\cref{tab:gen-quant} gives examples.

\begin{figure}
	\centering
	$\forall x\; \textit{picture}(x) \rightarrow \exists z \exists y\;
	\textit{tell}(y) \wedge \textit{story}(z) \wedge \textsc{arg1}(y,x) \wedge \textsc{arg2}(y,z)$%
	\vspace*{-1.5mm}
	\caption{%
		A first-order logical proposition,
		representing the most likely reading
		of \textit{Every picture tells a story}.
		Scope ambiguity is not discussed in this paper.
	}
	\label{fig:sem-comp-fol}
	
	\vskip .5\floatsep
	
	\begin{tikzpicture}[scopetree]
	\node (every) {$\textit{every}(x)$} ;
	\node [below left = of every] (picture) {$\textit{picture}(x)$} ;
	\node [below right = of every] (a) {$\textit{a}(z)$} ;
	\node [below left = of a] (story) {$\textit{story}(z)$} ;
	\node [below right = of a] (exists) {$\exists(y)$} ;
	\node [below left = of exists] (empty) {$\top$} ;
	\node [below right = of exists, align=center, yshift=-.5\baselineskip] (tell) {$\textit{tell}(y)\wedge\textsc{arg1}(y,x)$\\${}\wedge\textsc{arg2}(y,z)$} ;
	\draw (every.south) -- (picture.north) ;
	\draw (every.south) -- (a.north) ;
	\draw (a.south) -- (story.north) ;
	\draw (a.south) -- (exists.north) ;
	\draw (exists.south) -- (empty.north) ;
	\draw (exists.south) -- (tell.north) ;
	\end{tikzpicture}%
	\vspace*{-2mm}%
	\caption{%
		A scope tree, equivalent to \cref{fig:sem-comp-fol} above.
		Each non-terminal node is a quantifier, with its bound variable in brackets.
		Its left child is its restriction, and its right child its body.
	}%
	\label{fig:scope-tree}%
	
	\vskip .5\floatsep
	
	\begin{tabular}{cl}
		Quantifier & \multicolumn{1}{c}{Condition} \\
		\hline
		\it some & $|\R\cap\B| > 0$ \\
		\it every & $|\R\cap\B| = |\R|$ \\
		\it no & $|\R\cap\B| = 0$ \\
		\it most & $|\R\cap\B| > \frac{1}{2}|\R|$
	\end{tabular}
	\vspace*{-1.5mm}
	\captionof{table}{%
		Classical truth conditions for precise quantifiers,
		in generalised quantifier theory.
	}
	\label{tab:gen-quant}
	\vspace*{-2.5mm}
\end{figure}

\section{Generalised Quantifiers in Functional Distributional Semantics}
\label{sec:func-dist-sem}

Functional Distributional Semantics
defines a probabilistic graphical model for distributional semantics.
Importantly (from the point of view of formal semantics),
this graphical model incorporates a probabilistic version of model theory.

This is illustrated in \cref{fig:latent}.
The top row defines a distribution over situations,
each situation being an event with two participants.\footnote{%
	For situations with different structures
	(multiple events or different numbers of participants),
	we can define a family of such graphical models.
	Structuring the graphical model in terms of semantic roles
	makes the simplifying assumption
	that situation structure is isomorphic
	to a semantic dependency graph
	such as DMRS \citep{copestake2005mrs,copestake2009dmrs}.
	In the general case, the assumption fails.
	For example, the \textsc{arg3} of \textit{sell}
	corresponds to the \textsc{arg1} of \textit{buy}.
}
This generalises a model structure comprising a \textit{set} of situations,
as in classical situation semantics \citep{barwise1983situation}.
Each individual is represented by a \textit{pixie},
a point in a high-dimensional space,
which represents the features of the individual.
Two individuals could be represented by the same pixie,
and the space of pixies can be seen as a conceptual space
in the sense of \citet{gaerdenfors2000space,gaerdenfors2014space}.


The bottom row of the graphical model defines a distribution over truth values,
so that each predicate has some probability of being true of each individual.
Each predicate can therefore be seen as a probabilistic truth-conditional function.

\begin{figure}
	\centering
	\begin{tikzpicture}[default]
	
	\node[ent] (y) {$Y\!$} ;
	\node[ent, right=of y] (z) {$Z$} ;
	\node[ent, left=of y] (x) {$X$} ;
	
	\draw (y) -- (z) node[midway, above, color=black] {\textsc{arg2}} ;
	\draw (y) -- (x) node[midway, above, color=black] {\textsc{arg1}} ;
	
	\node[below right=3.5mm and 2.5mm of x, anchor=north west] {\textcolor{orange}{$\in \mathcal{X}$}} ;
	
	\draw (-1.5,-0.5) rectangle (1.5,-1.5) ;
	
	\node[val, below=of x] (tx) {$T_{r,\,X}$} ;
	\node[val, below=of y] (ty) {$T_{r,\,Y\!}$} ;
	\node[val, below=of z] (tz) {$T_{r,\,Z}$} ;
	
	\draw[dir] (x) -- (tx);
	\draw[dir] (y) -- (ty);
	\draw[dir] (z) -- (tz);
	
	\node[below right=3.5mm and 2.5mm of tx, anchor=north west] {\textcolor{lila}{$\in \cl{\bot,\top}$}} ;
	\node[xshift=-2ex, yshift=2ex] at (1.5, -1.5) {$\mathcal{V}$};
	\end{tikzpicture}
	\caption{%
		Probabilistic model theory, as formalised in Functional Distributional Semantics.
		Each node is a random variable.
		The plate (box in bottom row) denotes repetition of nodes.
		\newline
		\textbf{Top row:}
		pixie-valued random variables $X$, $Y$,~$Z$
		together represent a situation composed of three individuals.
		They are jointly distributed according to the semantic roles
		\textsc{arg1} and \textsc{arg2}.
		Their joint distribution can be seen as a probabilistic model structure.
		\newline
		\textbf{Bottom row:}
		each predicate~$r$ in the vocabulary~$\mathcal{V}$
		has a probabilistic truth-conditional function,
		which can be applied to each individual.
		This gives a truth-valued random variable
		for each individual for each predicate.
	}
	\label{fig:latent}
	\vspace*{-2mm}
\end{figure}

In this paper, I will not discuss learning such a model
\citex{for an up-to-date approach, see}{emerson2020pixie}.
Instead, the focus is on how we can manipulate a trained model,
to move from single predicates to complex propositions.

In previous work (\EC), I sketched an account of quantification.
The idea is to follow generalised quantifier theory,
but with a truth-valued random variable for each node in the scope tree.
Similarly to the classical case,
the distributions for these nodes are defined bottom-up through the tree.

In the classical theory, we only need to know the cardinalities
$|\R|$ and $|\R\cap\B|$.
In fact, all the conditions in \cref{tab:gen-quant}
can be expressed in terms of the ratio $\frac{|\R\cap\B|}{|\R|}$.
It therefore makes sense to consider the conditional probability $\Pc{b}{r,v}$,
because this uses the same ratio, as shown in~\cref{eqn:gen-quant}.\footnote{%
	I use uppercase for random variables,
	lowercase for values.
	I abbreviate $\Po{X\eq x}$ as $\Po{x}$,
	and $\Po{T\eq\top}$ as $\Po{t}$.
	For example, $\Pc{b}{r,v}$
	means $\Pc{B\eq\top}{R\eq\top,V\eq v}$.
}
\vspace*{-1mm}
\begin{equation}
\Pc{b}{r,v}
= \frac{\Pc{r,b}{v}}
{\Pc{r}{v}}
\label{eqn:gen-quant}
\end{equation}

More precisely,
$B$~and~$R$ are truth-valued random variables
for the body and restriction,
and $V$ is a tuple-of-pixies-valued random variable,
with one pixie for each free variable.
Intuitively, the truth of a quantified expression
depends on how likely $B$~is to be true, given that $R$~is true.\footnote{%
	This would not seem to cover so-called \textit{cardinal quantifiers}
	like \textit{one} and \textit{two}.
	Under \citet{link1983plural}'s lattice-theoretic approach,
	a model structure contains plural individuals,
	so numbers can be treated as normal predicates like adjectives.
	\label{fn:cardinal}
}

\begin{table}
	\centering
	\begin{tabular}{cc}
		Quantifier & Condition \\
		\hline
		\it some & $\Pc{b}{r,v} > 0$ \\
		\it every & $\Pc{b}{r,v} = 1$ \\
		\it no & $\Pc{b}{r,v} = 0$ \\
		\it most & $\Pc{b}{r,v} > \frac{1}{2}$
	\end{tabular}
	\caption{%
		Truth conditions for precise quantifiers,
		in terms of the conditional probability
		of the body given the restriction (and given all free variables).
		These conditions mirror \cref{tab:gen-quant}.
	}
	\label{tab:gen-quant-prob}
\end{table}

Truth conditions for quantifiers can be defined in terms of $\Pc{b}{r,v}$,
as shown in \cref{tab:gen-quant-prob}.
For these precise quantifiers,
the truth value is deterministic
-- if the condition in \cref{tab:gen-quant-prob} holds,
the quantifier's random variable $Q$ has probability~1 of being true,
otherwise it has probability~0.
However, taking a probabilistic approach means that we can naturally model
vague quantifiers like \textit{few} and \textit{many}.
I did not give further details on this point in \EC,
but I will expand on this in~\cref{sec:generic}.

\section{Quantification with Vague Predicates}
\label{sec:vague-quant}

Truth-conditional functions that give probabilities
strictly between 0 and~1
are motivated for both practical and theoretical reasons.
Practically, such a function can be implemented
as a feedforward neural network with a final sigmoid unit
(as used by \EC),
whose output is never exactly 0 or~1.
Theoretically, using intermediate probabilities of truth
allows a natural account of vagueness
\citep{goodman2015prob,sutton2015prob,sutton2017prob}.

However, as we will see in the following subsection,
intermediate probabilities pose a problem for \EC's account of quantification.

\subsection{Trivial Truth Values}
\label{sec:vague-quant:trivial}

Combining the conditions in \cref{tab:gen-quant-prob}
with vague predicates causes a problem,
which can be illustrated with a simple example.
Consider a model structure containing only a single individual,
and consider only the single predicate \textit{red},
which is true of this individual with probability~$p$.
Now consider the sentences \cref{ex:every,ex:some}.
\vspace*{-1mm}
\begin{exe}
	\ex\label{ex:every}
		Everything is red.
	\ex\label{ex:some}\vspace*{-1mm}
		Something is red.
\end{exe}
\vspace*{-1mm}

The body of each quantifier is simply the predicate \textit{red}.
For simplicity, we can assume that \textit{everything} and \textit{something}
put no constraints on their restrictions.
We need to calculate $\Pc{b}{r,v}$.
There are no free variables, and $R$ is always true,
so this is simply $\Po{b}$.
Because there is only one individual,
this is simply the probability~$p$.

This means that \cref{ex:every}~is true iff $p=1$,
and \cref{ex:some}~is false iff $p=0$.
However, we have seen above how predicates will never be true
with probability exactly 0 or exactly~1.
This means \cref{ex:every} is always false,
and \cref{ex:some} is always true,
even though we have assumed nothing about the individual!
\vspace*{-1mm}

\subsection{Distributions over Precise Predicates}
\label{sec:vague-quant:dist-prec}

To avoid the problem in~\cref{sec:vague-quant:trivial},
we must only combine precise quantifiers with precise predicates
(i.e.\ classical truth-conditional functions).
To do this, we can view a vague predicate
not as defining a probability of truth for each individual,
but as defining a distribution over precise predicates.
This induces a distribution for the quantifier.

Consider the example in~\cref{sec:vague-quant:trivial}.
With probability~$p$, \textit{red} is a precise predicate
that is true of the individual.
In this case, both \cref{ex:every} and~\cref{ex:some} are true.
With probability~$1\mn p$, \textit{red} is a precise predicate
that is false of the individual.
In this case, both \cref{ex:every} and~\cref{ex:some} are false.
Combining these cases,
both \cref{ex:every} and~\cref{ex:some} are true with probability~$p$,
which has avoided trivial truth values.

Formalising a vague predicate as a distribution over precise predicates
was also argued for by \citet{lassiter2011vague}.
It can be seen as an improved version of supervaluationism
(\citealp{fine1975vague,kamp1975vague};
\citealp[chapter 7]{keefe2000vague}),
which avoids the problem of higher-order vagueness,
as shown by \citeauthor{lassiter2011vague}.

\subsection{Probabilistic Scope Trees}
\label{sec:vague-quant:prob-tree}

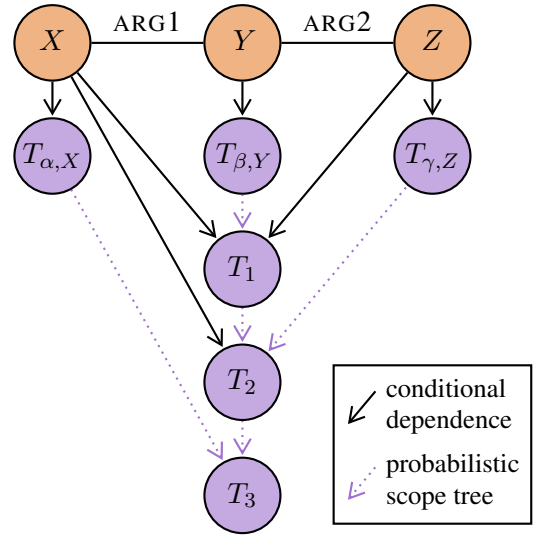
\begin{figure}
	\centering
	\begin{tikzpicture}[defshort, center coordinate=(y)]
	\node[ent] (y) {$Y\!$} ;
	\node[ent, right=of y] (z) {$Z$} ;
	\node[ent, left=of y] (x) {$X$} ;
	\draw (y) -- (z) node[midway, above] {\textsc{arg2}} ;
	\draw (y) -- (x) node[midway, above] {\textsc{arg1}} ;
	\node[val, below=of x] (tx) {$T_{\alpha,X}$} ;
	\node[val, below=of y] (ty) {$T_{\beta,Y\!}$} ;
	\node[val, below=of z] (tz) {$T_{\gamma,Z}$} ;
	\node[val, below=of ty] (t1) {$T_1$} ;
	\node[val, below=of t1] (t2) {$T_2$} ;
	\node[val, below=of t2] (t3) {$T_3$} ;
	\draw[dir] (x) -- (tx);
	\draw[dir] (y) -- (ty);
	\draw[dir] (z) -- (tz);
	\draw[dir] (x) -- (t1);
	\draw[dir] (z) -- (t1);
	\draw[dir] (x) -- (t2);
	\draw[scope] (t1) -- (ty);
	\draw[scope] (t2) -- (tz);
	\draw[scope] (t2) -- (t1);
	\draw[scope] (t3) -- (tx);
	\draw[scope] (t3) -- (t2);
	\draw[dir] (.7, -3.08) -- (.55, -3.38);
	\node[anchor=west, align=left, font=\captionfont] at (.75,-3.23)
		{conditional\\ dependence};
	\draw[scope] (.55, -4.05) -- (.7, -3.75);
	\node[anchor=west, align=left, font=\captionfont] at (.75,-3.9)
		{probabilistic\\ scope tree};
	\draw (.48,-2.85) rectangle (1.52,-4.25);
	\end{tikzpicture}%
	\vspace*{-1mm}%
	\caption{%
		A probabilistic scope tree.
		$T_1$,~$T_2$,~$T_3$ correspond to non-terminal nodes in \cref{fig:scope-tree},
		going up through the tree.
		One random variable is marginalised out at a time,
		until $T_3$ is no longer dependent on any variables.
	}%
	\label{fig:scope-graph}%
	\vspace*{-2mm}%
\end{figure}

To generalise the account in \cref{sec:vague-quant:dist-prec}
to arbitrary scope trees \seeref{sec:vague-quant:prob-tree-soft}
and vague quantifiers \seeref{sec:generic},
it is helpful to introduce a graphical notation
for \textit{probabilistic scope trees},
illustrated in \cref{fig:scope-graph}.
This makes the \EC account easier to visualise.
The improved proposal in this paper modifies
how the distribution for each truth value node is defined.

For a classical scope tree, the truth of a quantifier node
depends on its free variables,
and is defined in terms of the extensions of its restriction and body,
in a way that removes the bound variable.
For a probabilistic scope tree, the distribution for a quantifier node
is conditionally dependent on its free variables,
and is defined in terms of the distributions for its restriction and body,
marginalising out the bound variable.
The distributions at the leaves of the tree
are defined by predicates,
inducing a distribution for each quantifier node
as we work up through the tree.

\cref{fig:scope-graph} corresponds to \cref{fig:scope-tree},
if we set $\alpha$, $\beta$, $\gamma$ to be \textit{picture}, \textit{tell}, \textit{story}.
The distributions for $T_{\alpha,X}$, $T_{\beta,Y}$, $T_{\gamma,Z}$
are determined by the predicates.
We have three quantifier nodes in the classical scope tree,
and hence three additional truth value nodes in the probabilistic scope tree.
We first define a distribution for~$T_1$,
which represents the ${\exists(y)}$~quantifier,
and which depends on its free variables $X$~and~$Z$.
It is true if, for situations involving the fixed pixies $x$~and~$z$,
there is \textit{nonzero} probability that they are
the \textsc{arg1} and \textsc{arg2} of a telling-event pixie.
Next, we define a distribution for~$T_2$,
which represents the ${\textit{a}(z)}$~quantifier,
and depends on the free variable~$X$.
It is true if,
for situations involving the fixed pixie $x$ and story pixie $z$,
there is \textit{nonzero} probability that $T_1$ is true.
Finally, we define a distribution for~$T_3$,
which represents the ${\textit{every}(x)}$~quantifier,
and has no free variables.
It is true if,
for situations involving a picture pixie $X$,
we are \textit{certain} that $T_2$ is true.

\subsection{Probabilistic Scope Trees with Vague Predicates as Distributions}
\label{sec:vague-quant:prob-tree-soft}

\begin{figure*}
	\centering
	\begin{tikzpicture}[defwide]
	\node[ent] (y) {$Y\!$} ;
	\node[ent, right=of y] (z) {$Z$} ;
	\node[ent, left=of y] (x) {$X$} ;
	\draw (x) -- (y) node[midway, above] {\textsc{arg1}} ;
	\draw (y) -- (z) node[midway, above] {\textsc{arg2}} ;
	\node[val, below=of x] (tx) {$T_{\alpha,X}$} ;
	\node[val, below=of y] (ty) {$T_{\beta,Y\!}$} ;
	\node[val, below=of z] (tz) {$T_{\gamma,Z}$} ;
	\node[val, below=of ty] (t1) {$T_1$} ;
	\node[val, below=of t1] (t2) {$T_2$} ;
	\node[val, below=of t2] (t3) {$T_3$} ;
	\draw[dir] (x) -- (tx);
	\draw[dir] (y) -- (ty);
	\draw[dir] (z) -- (tz);
	\node[func, right=\funcsep of tx] (px) {$\Pi_\alpha$};
	\node[func, right=\funcsep of ty] (py) {$\Pi_\beta$};
	\node[func, right=\funcsep of tz] (pz) {$\Pi_\gamma$};
	\node[func, right=\funcsep of t1] (p1) {$\Pi_1$};
	\node[func, right=\funcsep of t2] (p2) {$\Pi_2$};
	\node[func, right=\funcsep of t3] (p3) {$\Pi_3$};
	\draw[dir] (px) -- (tx);
	\draw[dir] (py) -- (ty);
	\draw[dir] (pz) -- (tz);
	\draw[dir] (p1) -- (t1);
	\draw[dir] (p2) -- (t2);
	\draw[dir] (p3) -- (t3);
	\draw[dir, plila] (py) -- (p1);
	\draw[dir, plila] (p1) -- (p2);
	\draw[dir, plila] (p2) -- (p3);
	\draw[dir, plila] (pz) -- (p2);
	\draw[dir, plila] (px) to[out=-65, in=155] (p3);
	\draw[dir] (z) to[out=-145, in=30] (t1);
	\draw[dir] (x) to[out=-14, in=140, looseness=0.5] (t1);
	\draw[dir] (x) to[out=-21, in=130] (t2);
	\draw[dir] (.76, -3.1) -- (.64, -3.34);
	\node[anchor=west, align=left, font=\captionfont] at (.79,-3.25)
		{conditional dependence of\\ truth values};
	\draw[dir, plila] (.76, -3.7) -- (.64, -3.94);
	\node[anchor=west, align=left, font=\captionfont] at (.79,-3.85)
		{conditional dependence of\\ truth-conditional functions};
	\draw (.61,-2.87) rectangle (1.46,-4.2);
	\end{tikzpicture}%
	\vspace*{-1mm}%
	\caption{%
		The probabilistic scope tree in \cref{fig:scope-graph},
		explicitly showing random variables over precise functions.
	}%
	\vspace*{-2mm}%
	\label{fig:scope-graph-soft}%
\end{figure*}
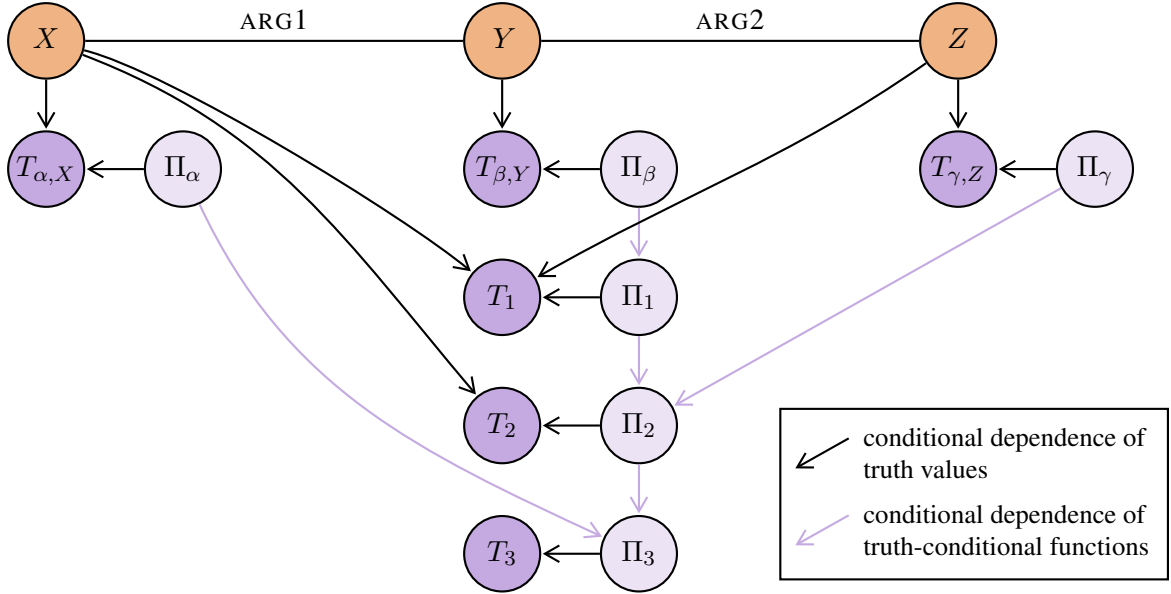

In this section
I show how to define the quantifier nodes in \cref{sec:vague-quant:prob-tree}
so that they are nontrivial.

To explicitly represent each vague truth-conditional function
as a random variable over precise functions,
we need to add a function node
for each truth value node in the graphical model.
For example, this transforms
\cref{fig:scope-graph} into \cref{fig:scope-graph-soft}.

For a truth value node~$T$ that is a leaf of the scope tree
(the second row of \cref{fig:scope-graph-soft}),
the distribution~$\Po{t}$ over truth values
follows the description in~\cref{sec:vague-quant:dist-prec}.
A precise predicate
${\pi:\mathcal{X}\rightarrow\{\top,\bot\}}$
maps pixies to truth values.
Given~$\pi$ and a pixie~$x$,
the distribution for $T$ is deterministic:
$T\eq\pi(x)$ with probability~1.
A distribution~$\Pi$ over precise predicates~$\pi$
defines a vague predicate~$\psi$,
by marginalising out this distribution:\footnote{%
	I write expectations with a subscript to indicate
	the random variable being marginalised out.
	To write the expectation in \cref{eqn:prec-sem-func}
	explicitly as a sum:
	$\Eo{\pi}{\pi(x)} = \sum_\pi[\pi(x)\Pr(\pi)]$.
}
\vspace*{-1mm}
\begin{equation}
\Pc{t}{x} = \psi(x) = \Eo{\pi}{\pi(x)}
\label{eqn:prec-sem-func}
\end{equation}

More generally, a truth value node~$Q$
is dependent on its free variables~$V$.
We can represent this in terms of a precise function
${\pi:\mathcal{X}^n\rightarrow\{\top,\bot\}}$,
where $n$ is the number of free variables.
Given values~$v$ for the free variables,
a distribution~$\Pi$ over precise predicates~$\pi$
defines a vague predicate~$\psi$,
by marginalising out this distribution:
\vspace*{-1mm}
\begin{equation}
\Pc{q}{v} = \psi(v) = \Eo{\pi}{\pi(v)}
\label{eqn:exp-prec-quant}
\end{equation}

What remains to be shown is that
the \EC account of quantification (in~\cref{sec:func-dist-sem})
can be adapted so that
a quantifier's distribution~$\Pi_Q$ over precise functions~$\pi_Q$
can be defined in terms of its restriction function~$\pi_R$
and body function~$\pi_B$.
This can be seen as probabilistic semantic composition:
the aim is to combine two truth-conditional functions
to produce a distribution over truth-conditional functions.
This is illustrated by the nodes $\Pi_1$, $\Pi_2$, $\Pi_3$ in \cref{fig:scope-graph-soft},
which are conditionally dependent on other function nodes
(indicated by the purple edges),
forming a probabilistic scope tree.

Expanding \cref{eqn:exp-prec-quant}
so it is dependent on the restriction and body functions,
we have~\cref{eqn:q-expQ}.
The aim is now to re-write the distribution for~$Q$,
using an adapted version of \EC,
in order to derive $\pi_Q$ in terms of $\pi_R$ and~$\pi_B$.
As explained in~\cref{sec:func-dist-sem},
the \EC account defines~$Q$
using the conditional probability $\Pc{b}{r,v}$.
More precisely, ${\Pc{q}{v}=f_Q(\Pc{b}{r,v})}$ for some~$f_Q$,
such as those defined by \cref{tab:gen-quant-prob}.
With vague functions now considered as
distributions over precise functions,
the conditional probability must be amended to
$\Pc{b}{r,v,\pi_R,\pi_B}$, as in~\cref{eqn:q-def},
given precise functions $\pi_R$~and~$\pi_B$
for the restriction and body.
This can be re-written as a ratio of probabilities
(corresponding to the classical sets),
summing over possible values for the bound variable(s)~$U$,
as in~\cref{eqn:q-sum}.
We can factorise out the distribution for $U$,
according to the conditional dependence structure
(illustrated in \cref{fig:scope-graph-soft}),
as in~\cref{eqn:q-fact}.
Finally, we can express $R$ and~$B$
in terms of the functions $\pi_R$ and~$\pi_B$,
and write the sum as an expectation,
as in~\cref{eqn:q-rat-exp}.
Note that $\pi_R$ and~$\pi_B$ take ${u\cup v}$ as an argument
-- by definition of a scope tree,
if we combine a quantifier's bound and free variables,
we get the free variables of its restriction and body.
I have written ${u\cup v}$ rather than ${\{u\}\cup v}$,
to leave open the possibility
that the quantifier has more than one bound variable,
which will be relevant in~\cref{sec:generic}.
\vspace*{-1mm}
\begin{align}
&\Pc{q}{v,\pi_R,\pi_B}
=\Eo{\pi_Q|\pi_R,\pi_B}{\pi_Q(v)}
\label{eqn:q-expQ}
\\
&= f_Q\rd{\big.
	\Pc{b}{r,v,\pi_R,\pi_B}
}
\label{eqn:q-def}
\\
&= f_Q\rd{
	\frac{
		\sum_u\Pc{b,r,u}{v,\pi_R,\pi_B}
	}{
		\sum_u\Pc{r,u}{v,\pi_R,\pi_B}
	}
}
\label{eqn:q-sum}
\\
&= f_Q\rd{
	\frac{
		\sum_u\Pc{u}{v}\Pc{r,b}{u,v,\pi_R,\pi_B}
	}{
		\sum_u\Pc{u}{v}\Pc{r}{u,v,\pi_R}
	}
}
\label{eqn:q-fact}
\\
&= f_Q\rd{
	\frac{
		\Eo{u|v\!}{\big.\pi_R(u\cup v)\,\pi_B(u\cup v)}
	}{
		\Eo{u|v\!}{\big.\pi_R(u\cup v)}
	}
}
\label{eqn:q-rat-exp}
\end{align}
\vspace*{-2mm}

\cref{eqn:q-rat-exp} gives a probability of truth,
hence a vague function.
Viewing it as a distribution over precise functions
(as in~\cref{sec:vague-quant:dist-prec}),
we finally have a definition of $\pi_Q$
in terms of $\pi_R$ and~$\pi_B$.
Concretely, $\pi_Q$ returns truth
iff \cref{eqn:q-rat-exp} is above a threshold.
A uniform distribution over thresholds in~$[0,1]$
gives a distribution over such functions.

Abbreviating the notation, we can write~\cref{eqn:q-summary}.
A quantifier's truth-conditional function
depends on the restriction and body functions,
marginalising out the bound variable.
The ratio of expectations mirrors
the classical ratio of cardinalities.
\begin{equation}
\pi_Q \sim f_Q\rd{
	\frac{
		\Eo{u}{\pi_R \pi_B}
	}{
		\Eo{u}{\pi_R}
	}
}
\label{eqn:q-summary}
\end{equation}

We can now recursively define functions for quantifier nodes,
given functions in the leaves.
We can therefore see \cref{fig:scope-graph}
as an abbreviated notation for \cref{fig:scope-graph-soft}.
The dotted edges do not indicate
conditional dependence of \textit{truth values},
but conditional dependence of \textit{truth-conditional functions}.

\section{Vague Quantifiers and Generics}
\label{sec:generic}

While \textit{some}, \textit{every}, \textit{no}, and \textit{most}
can be given precise truth conditions,
other natural language quantifiers are vague.
In particular, we can consider the terms \textit{few} and \textit{many}.\footnote{%
	\citet{partee1988many} surveys work suggesting that
	\textit{few} and \textit{many} are ambiguous
	between a vague cardinal reading and a vague proportional reading.
	As mentioned in \cref{fn:cardinal},
	we can treat cardinals as predicates rather than quantifiers.
}

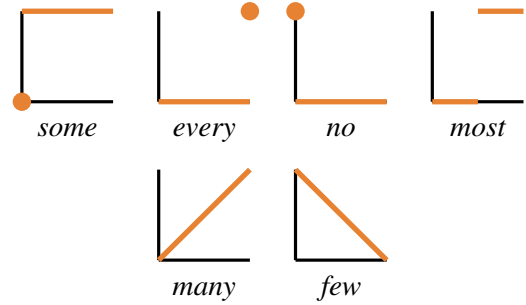
\begin{figure}
	\centering
	\begin{tikzpicture}[very thick, x=12mm, y=12mm]
	\tikzstyle{curve}=[orange, line width=2pt]
	\tikzstyle{label}=[anchor=north, inner sep=1mm]
	\def\y{1.75}
	
	\draw (0,\y) -- +(0,1);
	\draw (0,\y) -- +(1,0);
	\node[label] at (0.5,\y) {\it some};
	\draw[orange, fill=orange] (0,\y) circle[radius=1mm];
	\draw[curve] (0,\y+1) -- +(1,0);
	
	\draw (1.5,\y) -- +(0,1);
	\draw (1.5,\y) -- +(1,0);
	\node[label] at (2,\y) {\it every};
	\draw[orange, fill=orange] (2.5,\y+1) circle[radius=1mm];
	\draw[curve] (1.5,\y) -- +(1,0);
	
	\draw (3,\y) -- +(0,1);
	\draw (3,\y) -- +(1,0);
	\node[label] at (3.5,\y) {\it no};
	\draw[orange, fill=orange] (3,\y+1) circle[radius=1mm];
	\draw[curve] (3,\y) -- +(1,0);
	
	\draw (4.5,\y) -- +(0,1);
	\draw (4.5,\y) -- +(1,0);
	\node[label] at (5,\y) {\it most};
	\draw[curve] (4.5,\y) -- ++(.5,0) ++(0,1) -- ++(.5,0);
	
	\draw (1.5,0) -- +(0,1);
	\draw (1.5,0) -- +(1,0);
	\node[label] at (2,0) {\it many};
	\draw[curve] (1.5,0) -- (2.5,1);
	
	\draw (3,0) -- +(0,1);
	\draw (3,0) -- +(1,0);
	\node[label] at (3.5,0) {\it few};
	\draw[curve] (3,1) -- (4,0);
	
	\end{tikzpicture}%
	\vspace*{-2mm}%
	\caption{%
		Probabilities of truth for various quantifiers.
		Each x-axis is $\Pc{b}{r,v,\pi_R,\pi_B}$,
		and each y-axis is $\Pc{q}{v,\pi_R,\pi_B}$,
		plotting the function~$f_Q$ in orange.
		All axes range from 0~to~1.
		Quantifiers in the bottom row are vague,
		requiring intermediate probabilities.
	}%
	\vspace*{-1mm}%
	\label{fig:quant}%
\end{figure}

Under a classical account \citeg{barwise1981quant},
\textit{many} means that ${\R\cap\B}$ is large compared to $\R$,
but how large is underspecified;
similarly, \textit{few} means this ratio is small.
The underspecification of a proportion
can naturally be represented as a distribution.
So, we can define the meaning of a vague generalised quantifier
to be a function from $\Pc{b}{r,v}$ to a probability of truth,
as illustrated in \cref{fig:quant}.

A particularly challenging case of natural language quantification
involves \textit{generic} sentences,
such as: \textit{dogs bark},
\textit{ducks lay eggs},
and \textit{mosquitoes carry malaria}.
Generics are ubiquitous in natural language,
but they are challenging for classical models,
because the truth conditions seem to depend heavily
on lexical semantics and on the context of use
\citex{for discussion, see}{carlson1977generic,carlson1995generic,leslie2008generic}.

While it is tempting to treat generic quantification
as underspecification of a precise quantifier
\citeg{herbelot2010quant,herbelot2011underquant},
this is at odds with evidence that generics
are easier for children to acquire than precise quantifiers
\citep{hollander2002generic,leslie2008generic,gelman2015generic},
and also easier for adults to process
\citep{khemlani2007generic}.

In contrast, \citet{tessler2019generic} analyse generic sentences
as being semantically simple,
with the complexity coming down to pragmatic inference.
\nocite{tessler2018thesis}
They use Rational Speech Acts (RSA), a Bayesian approach to pragmatics
\citep{frank2012rsa,goodman2016rsa}.
In this framework, literal truth is separated from pragmatic meaning.
Communication is viewed as a game
where a listener has a prior belief about a situation,
and a speaker wants to update the listener's belief.
Given a truth-conditional function,
a \textit{literal listener} updates their belief by conditioning on truth,
ruling out situations for which the function returns false.
A \textit{pragmatic speaker} who observes a situation
can choose an utterance which is informative for a literal listener
-- in particular, the utterance which maximises
a literal listener's posterior probability for the observed situation.
A \textit{pragmatic listener} can update their belief
by conditioning on a pragmatic speaker's utterance.

\citeauthor{tessler2019generic}'s insight
is that this inference of \textit{pragmatic} meanings
can account for the behaviour of generic sentences.
The literal meaning of a generic can be simple
(it is more likely to be true as the proportion increases),
but the pragmatic meaning can have a rich dependence
on the world knowledge encoded in the prior over situations.
For example, \textit{Mosquitoes carry malaria}
does not mean that all mosquitoes do
(in fact, many do not)
but it can be informative for the listener:
as most animals never carry malaria,
even a small proportion is pragmatically relevant.

Building on this,
we could model the generic quantifier
by setting $f_Q$ as the identity function
(the same as \textit{many} in \cref{fig:quant}).
From~\cref{eqn:q-rat-exp},
the probability of truth is then
as shown in~\cref{eqn:gen-def-1}.
However, marginalising out $\Pi_R$ and~$\Pi_B$
is computationally expensive,
as it requires summing over all possible functions.
We can approximate this by reversing the order of the expectations,
and so marginalising out $\Pi_R$ and~$\Pi_B$ before~$U$,
as shown in~\cref{eqn:gen-def-2},
where $\psi_R$~and~$\psi_B$ are vague functions.
Evaluating a vague function is computationally simple.
\vspace*{-1mm}
\begin{align}
&\Eo{\pi_R,\pi_B\!}{
	\frac{
		\Eo{u|v}{\big. \pi_R(u\cup v)\,\pi_B(u\cup v)}
	}{
		\Eo{u|v}{\big. \pi_R(u\cup v)}
	}
}
\label{eqn:gen-def-1}
\\
&\approx
\frac{
	\Eo{u|v}{\big. \psi_R(u\cup v)\,\psi_B(u\cup v)}
}{
	\Eo{u|v}{\big. \psi_R(u\cup v)}
}
\label{eqn:gen-def-2}
\end{align}
\vspace*{-2mm}

Abbreviating this, similarly to~\cref{eqn:q-summary},
we can write:
\vspace*{-1mm}
\begin{equation}
\psi_Q = \frac{
	\Eo{u}{\psi_R \psi_B}
}{
	\Eo{u}{\psi_R}
}
\label{eqn:gen-summary}
\end{equation}
\vspace*{-2mm}

For precise quantifiers,
using vague functions gives trivial truth values
(discussed in~\cref{sec:vague-quant:trivial}),
but for generics,
\cref{eqn:gen-def-1} and~\cref{eqn:gen-def-2}
give similar probabilities of truth.
To put it another way,
a vague quantifier doesn't need precise functions.
Modelling generics with~\cref{eqn:gen-def-1}
was driven by the intuition
that generics are vague but semantically simple.
The alternative in~\cref{eqn:gen-def-2} is even simpler,
because we only need to calculate~$\E{u|v}$ once in total,
rather than once for each possible $\pi_R$~and~$\pi_B$.
This would make generics computationally simpler than other quantifiers,
consistent with the evidence that they are easier to acquire and to process.


\begin{figure}
	\centering%
	\begin{tikzpicture}[defcompact]
		\node[ent] (y) {$Y\!$} ;
		\node[ent, right=of y] (z) {$Z$} ;
		\node[ent, left=of y] (x) {$X$} ;
		\draw (y) -- (z) node[midway, above] {\textsc{arg2}} ;
		\draw (y) -- (x) node[midway, above] {\textsc{arg1}} ;
		\node[val, below=of x] (tx) {$T_{\alpha,X}$} ;
		\node[val, below=of y] (ty) {$T_{\beta,Y\!}$} ;
		\node[val, below=of z] (tz) {$T_{\gamma,Z}$} ;
		\node[val, below=of x, xshift=-\scopesep] (td) {$T_{\delta,X}$} ;
		\draw[dir] (x) -- (tx);
		\draw[dir] (y) -- (ty);
		\draw[dir] (z) -- (tz);
		\draw[dir] (x) -- (td);
		\node[val, below=of ty] (t1) {$R$} ;
		\node[val, below=of t1] (t2) {$Q$} ;
		\draw[dir] (tx) -- (t1);
		\draw[dir] (ty) -- (t1);
		\draw[dir] (tz) -- (t1);
		\draw[scope] (t2) -- (t1);
		\draw[scope] (t2) -- (td);
	\end{tikzpicture}%
	\vspace*{-.5mm}
	\caption{%
		\citet{emerson2017a}'s logical inference,
		re-analysed as generic quantification.
		$R$ is the restriction, the logical conjunction of 
		$T_{\alpha,X}$,
		$T_{\beta,Y}$,
		and $T_{\gamma,Z}$,
		while $T_{\delta,X}$ is the body.
		Generic quantification gives
		$\Pr(q)=\Pc{t_{\delta,X}}{t_{\alpha,X},t_{\beta,Y},t_{\gamma,Z}}$,
		marginalising out all three bound variables ($X$, $Y$, and $Z$).
	}
	\label{fig:gen-three}
\end{figure}
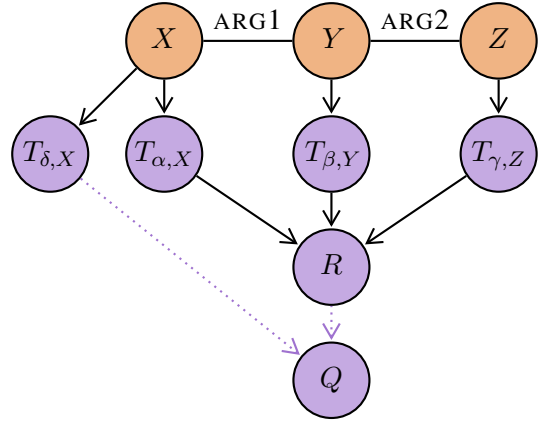

In fact, \cref{eqn:gen-def-2} takes us back to
\EC's conditional probability,
as shown in~\cref{eqn:gen-def-prob}.
\vspace*{-1mm}
\begin{align}
\psi_Q(v) &= \frac{
	\sum_u \Pc{u}{v}\Pc{r}{u,v}\Pc{b}{u,v}
}{
	\sum_u \Pc{u}{v}\Pc{r}{u,v}
}
\notag
\\
&= \Pc{b}{r,v}
\label{eqn:gen-def-prob}
\end{align}

This means the logical inference
proposed by \citet{emerson2017a}
can in fact be seen as generic quantification.
This is illustrated in \cref{fig:gen-three},
which corresponds to a sentence like
\textit{Rooms that have stoves are kitchens},
if $\alpha$,~$\beta$,~$\gamma$,~$\delta$ are set to
\textit{room}, \textit{have}, \textit{stove}, \textit{kitchen}.\footnote{%
	An example from RELPRON \citep{rimell2016relpron}.
}

Not only does this approach to quantification
deal with both precise and vague quantifiers in a uniform way,
it can also explain why generics are easier to process than precise quantifiers.

\section{Donkey Sentences}
\label{sec:donkey}

An example of a donkey sentence is shown in~\cref{ex:every-donkey}.
They are challenging for classical semantic theories,
because naive composition, shown in~\cref{ex:every-donkey-unbound},
leaves a variable~($y$) outside the scope of its quantifier
\citep{geach1962donkey}.
The tempting solution in~\cref{ex:every-donkey-strong}
requires a universal quantifier for an indefinite (\textit{a~donkey}),
which would be non-compositional.\footnote{%
	For simplicity,
	\cref{ex:every-donkey-unbound,ex:every-donkey-strong}
	suppress event variables.
}
\begin{exe}
	\ex Every farmer who owns a donkey feeds it.%
	\label{ex:every-donkey}
	\ex $\forall x\big[\big(\textit{farmer}(x)
		\wedge\exists y[\textit{donkey}(y)\wedge\newline\textit{own}(x,y)]\big)
		\rightarrow\textit{feed}(x,y)\big]$
	\label{ex:every-donkey-unbound}
	\ex $\forall x\forall y\big[\big(\textit{farmer}(x)
	\wedge\textit{donkey}(y)\wedge\newline\textit{own}(x,y)\big)
	\rightarrow\textit{feed}(x,y)\big]$
	\label{ex:every-donkey-strong}
\end{exe}

\citet{kanazawa1994donkey},
\citet{brasoveanu2008donkey},
and \citet{king2016anaphora}
discuss how donkey sentences seem to admit multiple readings,
which vary in the strength of their truth conditions,
and which depend on both lexical semantics
and the context of use.
This kind of dependence is exactly what
\citet{tessler2019generic} explained using RSA,
so I will apply the same tools here.

As discussed in~\cref{sec:generic},
generics are more basic than classical quantifiers,
so I first consider generic donkey sentences,
as illustrated in \cref{ex:donkey,ex:linguist,ex:mosquito}.
An analysis of \cref{ex:every-donkey} is given in \cref{sec:appendix}.
\vspace*{-1mm}
\begin{exe}
	\ex Farmers who own donkeys feed them.
	\label{ex:donkey}\vspace*{-1mm}
	\ex Linguists who use probabilistic models love them.
	\label{ex:linguist}\vspace*{-1mm}
	\ex Mosquitoes which bite birds infect them with malaria.
	\label{ex:mosquito}\vspace*{-1mm}
\end{exe}

Example \cref{ex:mosquito} shows
it is inappropriate to use a universal quantifier:
not all mosquitoes carry malaria,
and not all bitten birds are infected
(even if bitten by a malaria-carrying mosquito).
However, this sentence still communicates
that malaria is spread between birds by mosquitoes.
This relies on pragmatic inference,
from prior knowledge that most animals cannot spread malaria.

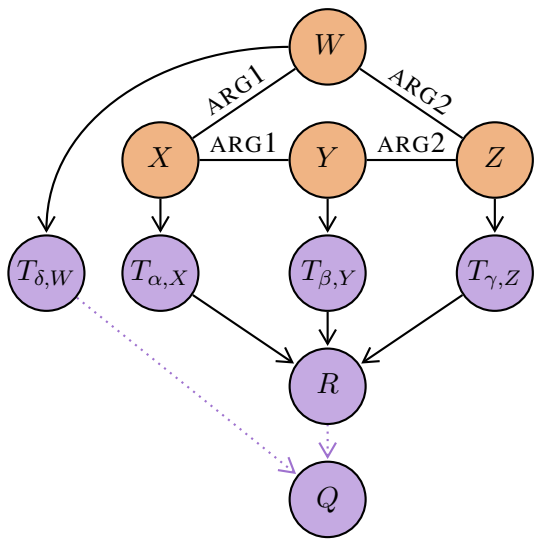
\begin{figure}
	\centering%
	\begin{tikzpicture}[defcompact]
	\node[ent] (w) {$W\!$} ;
	\node[ent, below=of w] (y) {$Y$} ;
	\node[ent, right=of y] (z) {$Z$} ;
	\node[ent, left=of y] (x) {$X$} ;
	\draw (y) -- (z) node[midway, above, yshift=-.5mm] {\textsc{arg2}} ;
	\draw (y) -- (x) node[midway, above, yshift=-.5mm] {\textsc{arg1}} ;
	\draw (w) -- (z) node[midway, above, sloped, yshift=-.5mm] {\textsc{arg2}} ;
	\draw (w) -- (x) node[midway, above, sloped, yshift=-.5mm] {\textsc{arg1}} ;
	\node[val, below=of x] (tx) {$T_{\alpha,X}$} ;
	\node[val, below=of y] (ty) {$T_{\beta,Y\!}$} ;
	\node[val, below=of z] (tz) {$T_{\gamma,Z}$} ;
	\node[val, below=of x, xshift=-\scopesep] (td) {$T_{\delta,W}$} ;
	\draw[dir] (x) -- (tx);
	\draw[dir] (y) -- (ty);
	\draw[dir] (z) -- (tz);
	\draw[dir] (w) to[out=180,in=90] (td);
	\node[val, below=of ty] (t1) {$R$} ;
	\node[val, below=of t1] (t2) {$Q$} ;
	\draw[dir] (tx) -- (t1);
	\draw[dir] (ty) -- (t1);
	\draw[dir] (tz) -- (t1);
	\draw[scope] (t2) -- (t1);
	\draw[scope] (t2) -- (td);
	\end{tikzpicture}%
	\vspace*{-1mm}%
	\caption{%
		Analysis of a generic donkey sentence,
		using generic quantification.
		The quantifier node $Q$ has
		restriction $R$ (a logical conjunction)
		and body $T_{\delta,W}$.
	}
	\label{fig:donkey}
\end{figure}

Despite the challenge for classical theories, generic donkey sentences
can be straightforwardly handled by my proposed probabilistic approach.
An example is shown in \cref{fig:donkey},
which corresponds to~\cref{ex:donkey},
if $\alpha$, $\beta$, $\gamma$, $\delta$ are set to
\textit{farmer}, \textit{own}, \textit{donkey}, \textit{feed}.
Intuitively, the more likely it is that
a farmer owning a donkey implies the farmer feeding the donkey,
the more likely it is for the sentence to be true.
Given world knowledge and a discourse context,
this can lead to a sharp threshold for being uttered,
using RSA's pragmatic inference.

\section{Related Work}

Functional Distributional Semantics
is related to other probabilistic semantic approaches.
\citet{goodman2015prob} and \citet{bernardy2018bayes,bernardy2019bayes}
represent meaning as a probabilistic program.
This paper brings Functional Distributional Semantics
closer to their work,
because a probabilistic scope tree can be seen as a probabilistic program.
An important practical difference
is that Functional Distributional Semantics
represents all predicates in the same way (as functions of pixies),
allowing a model to be trained on corpus data.

Probabilistic TTR \citep{cooper2005type,cooper2015prob}
also represents meaning as a probabilistic truth-conditional function.
However, in this paper I have provided an alternative compositional semantics,
in order to deal with vague quantifiers and generics.
In principle, my proposal could be incorporated into a probabilistic TTR appproach.
Furthermore, although \citet{cooper2015prob} discuss learning,
they assume a richer input than available in distributional semantics.


Some hybrid distributional-logical systems exist
\citeg{lewis2013logic,grefenstette2013tensor,herbelot2015quantifier,beltagy2016logic},
but these do not discuss challenging cases
like generics and donkey sentences.

Explaining the multiple readings of donkey sentences using pragmatic inference
has been proposed using non-probabilistic tools
\citeg{champollion2016homogeneity,champollion2019donkeys}.
I have provided a concrete computational method to calculate such inferences,
in the same way that \citet{tessler2019generic}
have provided a concrete account of generics.

\section{Conclusion}

In this paper, I have presented a compositional semantics
for both precise and vague quantifiers,
in the probabilistic framework of Functional Distributional Semantics.
I have re-interpreted previous work in this framework
as performing generic quantification,
building on the approach of \citet{tessler2019generic}.
I have shown how generic quantification is
computationally simpler than classical quantification,
consistent with evidence that generics are a ``default'' mode of processing.
Finally, I have presented examples of generic donkey sentences,
which are doubly challenging for classical theories,
but straightforward under my proposed approach.

\section*{Acknowledgements}

This paper builds on chapter 7 of my PhD thesis
\citep{emerson2018},
and I would like to thank my PhD supervisor Ann Copestake,
for her support, advice, and suggestions.
I would also like to thank the anonymous reviewers,
for pointing out areas that were unclear,
and suggesting additional areas for discussion.

I am supported by a Research Fellowship
at Gonville \& Caius College, Cambridge.

\bibliography{thesis,pixie,generic}
\bibliographystyle{acl_natbib}

\begin{figure*}
\centering
\begin{tikzpicture}[defcompact]
	\node[ent] (w) {$W\!$} ;
	\node[ent, below=of w] (y) {$Y$} ;
	\node[ent, right=of y] (z) {$Z$} ;
	\node[ent, left=of y] (x) {$X$} ;
	\draw (y) -- (z) node[midway, above, yshift=-.5mm] {\textsc{arg2}} ;
	\draw (y) -- (x) node[midway, above, yshift=-.5mm] {\textsc{arg1}} ;
	\draw (w) -- (z) node[midway, above, sloped, yshift=-.5mm] {\textsc{arg2}} ;
	\draw (w) -- (x) node[midway, above, sloped, yshift=-.5mm] {\textsc{arg1}} ;
	\node[val, below=of x] (tx) {$T_{\alpha,X}$} ;
	\node[val, below=of y] (ty) {$T_{\beta,Y\!}$} ;
	\node[val, below=of z] (tz) {$T_{\gamma,Z}$} ;
	\node[val, below right=of z] (td) {$T_{\delta,W}$} ;
	\draw[dir] (x) -- (tx);
	\draw[dir] (y) -- (ty);
	\draw[dir] (z) -- (tz);
	\draw[dir] (w) to[out=0,in=90] (td);
	\node[val, below=of ty] (t1) {$Q_{\mathrm{E1}}$} ;
	\node[val, below=of t1] (t2) {$T_{\mathrm{RC}}$} ;
	\node[val, below=of t2] (t3) {$Q_\exists$} ;
	\node[val, below right=.5 and 1 of t2] (t4) {$T_{\mathrm{DP}}$} ;
	\node[val, below=of td] (t5) {$Q_{\mathrm{E2}}$} ;
	\node[val, below right=.5 and 1 of t4] (t6) {
		$\!Q_{\mathrm{\mspace{-2mu}G\mspace{-1.5mu}E\mspace{-1mu}N}}\!$
	} ;
	\node[val, below=of t4] (t7) {$Q_\forall$} ;
	\draw[dir] (x) -- (t1);
	\draw[dir] (z) -- (t1);
	\draw[scope] (t1) -- (ty);
	\draw[dir] (t1) -- (t2);
	\draw[dir] (tx) -- (t2);
	\draw[scope] (t3) -- (t2);
	\draw[scope] (t3) -- (tz);
	\draw[dir] (x) to[out=-135, in=135] (t3);
	\draw[dir] (t2) -- (t4);
	\draw[dir] (tz) -- (t4);
	\draw[scope] (t5) -- (td);
	\draw[dir] (z) to[out=-45, in=115] (t5);
	\draw[dir] (x) to[out=-30, in=170, looseness=.8] (1,-1.4)
		to[out=-10, in=130, looseness=.7] (t5);
	\draw[scope] (t6) -- (t5);
	\draw[scope] (t6) -- (t4);
	\draw[dir] (x) to[out=-35, in=170, looseness=.8] (.95,-1.45)
		to[out=-10, in=110, looseness=.7] (t6);
	\draw[scope] (t7) -- (t3);
	\draw[scope] (t7) -- (t6);
\end{tikzpicture}%
\vspace*{-.5mm}
\end{figure*}

\newpage
\strut
\newpage
\appendix

\section{Classical Donkey Sentences}
\label{sec:appendix}

In this analysis of a classical donkey sentence,
the donkey pronoun is associated with a generic quantifier,
while all other quantifiers are precise.
The generic quantifier allows the range of readings
associated with donkey sentences.

The above figure corresponds to example~\cref{ex:every-donkey},
if $\alpha$, $\beta$, $\gamma$, $\delta$ are set to
\textit{farmer}, \textit{own}, \textit{donkey}, \textit{feed}.
Intuitively, this analysis says that,
if all farmers who own at least one donkey
feed at least a proportion~$p$ of their donkeys,
then this sentence is true with probability~$p$.

The probability of truth gradually increases with the proportion~$p$.
Given world knowledge and a discourse context,
this can lead to a sharp threshold proportion
above which it is uttered,
using pragmatic inference in the RSA framework.
If distinguishing small proportions is pragmatically relevant,
the \textit{weak} reading becomes preferred.
If distinguishing large proportions is pragmatically relevant,
the \textit{strong} reading becomes preferred.

I will now go over all nodes in the graph.
Firstly, the distributions for
$T_{\alpha,X}$, $T_{\beta,Y}$, $T_{\gamma,Z}$, $T_{\delta,W}$
are determined by the predicates.

The remaining truth value nodes
are labelled for convenience.
$T_{\mathrm{RC}}$ and $T_{\mathrm{DP}}$ are logical conjunctions
(for the relative clause and donkey pronoun, respectively).
The remaining five nodes are quantifier nodes,
each quantifying one variable.

Note that $Z$ is quantified twice
(by $Q_\exists$ and $Q_{\mathrm{GEN}}$).
This would be surprising in a classical logic,
but is not a problem here
-- marginalising out a random variable
means that the quantifier node is not dependent on that variable,
but the random variable is still part of the joint distribution,
so it can be referred to by other nodes.
Because of this double quantification, the scope ``tree''
is actually a scope DAG (directed acyclic graph).

$Q_{\mathrm{E1}}$ and $Q_{\mathrm{E2}}$
marginalise out the event variables, respectively $Y$ and~$W$,
with trivially true restrictions
and bodies $T_{\beta,Y}$ and~$T_{\delta,W}$,
leaving free variables $X$ and~$Z$.
They can be treated like \textit{some} in \cref{fig:quant}.
For given pixies $x$ and $z$,
$Q_{\mathrm{E1}}$ is true if $x$ owns $z$;
$Q_{\mathrm{E2}}$ is true if $x$ feeds $z$.

$Q_\exists$ marginalises out~$Z$,
with $T_{\mathrm{RC}}$ as restriction and $Q_{\mathrm{E1}}$ as body,
leaving the free variable~$X$.
It can be treated like \textit{some} in \cref{fig:quant}.
For a given~$x$,
it is true if $x$ is a farmer
and there is a donkey~$z$ such that $Q_{\mathrm{E1}}$ is true.

$Q_{\mathrm{GEN}}$ also marginalises out~$Z$,
with $T_{\textrm{DP}}$ as restriction and $Q_{\mathrm{E2}}$ as body,
leaving the free variable~$X$.
It uses the generic quantifier, as in~\cref{eqn:gen-def-2}.
For a given~$x$,
it considers donkeys~$z$ for which $Q_{\mathrm{RC}}$ is true;
the probability of truth
is the proportion of such~$z$ for which $Q_{\mathrm{E2}}$ is true
(out of donkeys owned by farmer~$x$,
the proportion fed by~$x$).

Finally, $Q_\forall$ marginalises out $X$,
with $T_\exists$ as restriction and $Q_{\textrm{GEN}}$ as body,
leaving no free variables.
It is treated as in \cref{fig:quant}.
It is true if,
whenever $T_\exists$ is true of $x$,
$Q_{\textrm{GEN}}$ is true of $x$,
considering $Q_{\textrm{GEN}}$ as a distribution over precise functions.

\end{document}